\documentclass{article}

\usepackage{PRIMEarxiv}

\usepackage[utf8]{inputenc} 
\usepackage[T1]{fontenc}    
\usepackage{hyperref}       
\usepackage{url}            
\usepackage{booktabs}       
\usepackage{amsfonts}       
\usepackage{nicefrac}       
\usepackage{microtype}      
\usepackage{lipsum}
\usepackage{fancyhdr}       
\usepackage{graphicx}       
\usepackage{orcidlink}
\usepackage{amsmath}
\usepackage{cleveref}

\usepackage{array}
\newcolumntype{C}[1]{>{\centering\arraybackslash}p{#1}}

\graphicspath{{media/}}     

\title{Context-Aware Full Body Anonymization using Text-to-Image Diffusion Models
}

\author{
    Pascal Zwick \\
    FZI Research Center for Information Technology \\
    76131 Karlsruhe, Germany \\
    \textit{zwick@fzi.de} \\
    \And
    Kevin Roesch \\
    FZI Research Center for Information Technology \\
    76131 Karlsruhe, Germany \\
    \textit{kevin.roesch@fzi.de} \\
    \And
    Marvin Klemp \\
    Karlsruhe Institute of Technology \\
    76131 Karlsruhe, Germany \\
    \texttt{marvin.klemp@kit.edu} \\
    \And
    Oliver Bringmann \\
    FZI Research Center for Information Technology \\
    University of Tuebingen \\
    72074 Tuebingen, Germany \\
    \textit{oliver.bringmann@uni-tuebingen.de} \\
}

\begin{document}






\maketitle

\vspace{-40pt}

\begin{figure}[h]
    \centering
    \includegraphics[width=0.7\textwidth]{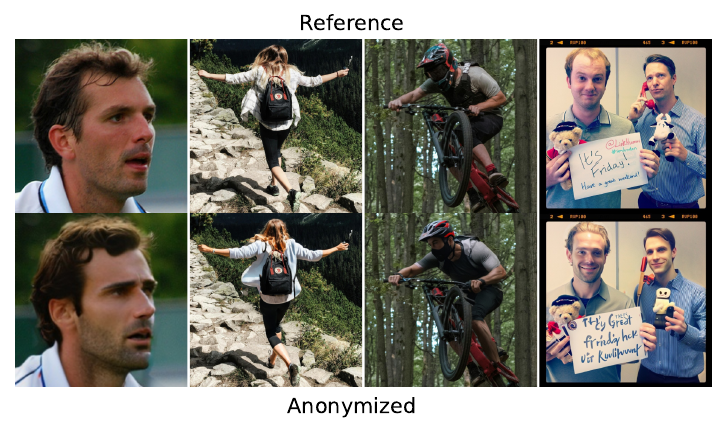}
    \caption{Images from different scenes anonymized with the method proposed in this paper. \cite{lin2014microsoft}}
    \label{fig:title}
\end{figure}

\begin{abstract}

Anonymization plays a key role in protecting sensible information of individuals in real world datasets. Self-driving cars for example need high resolution facial features to track people and their viewing direction to predict future behaviour and react accordingly. In order to protect people's privacy whilst keeping important features in the dataset, it is important to replace the full body of a person with a highly detailed anonymized one. In contrast to doing face anonymization, full body replacement decreases the ability of recognizing people by their hairstyle or clothes. In this paper, we propose a workflow for full body person anonymization utilizing Stable Diffusion as a generative backend. Text-to-image diffusion models, like Stable Diffusion, OpenAI's DALL-E or Midjourney, have become very popular in recent time, being able to create photorealistic images from a single text prompt. We show that our method outperforms state-of-the art anonymization pipelines with respect to image quality, resolution, Inception Score (IS) and Frechet Inception Distance (FID). Additionally, our method is invariant with respect to the image generator and thus able to be used with the latest models available.

\end{abstract}

\section{Introduction}
Image based deep learning models require a lot of high quality images for training, be it for classification, movement prediction or other problems. In order to not infringe the privacy of individuals, training data needs to be anonymized, leading to less data with a reduction in quality. As an example, the detection and viewing direction of pedestrians is needed for movement prediction in self driving car applications.

With the rise of text-to-image models, like Stable Diffusion \cite{Rombach_2022_CVPR}, DALL-E \cite{ramesh2021zero} or Midjourney \cite{midjourney}, it is now possible to generate realistic looking images from text prompts. Although so-called diffusion models are compute intensive compared to previous methods, like GANs, there exist a lot of pretrained models with different characteristics and output resolution, up to $1024 \times 1024$ for Stable Diffusion XL \cite{podell2023sdxl}. However, resolution is often not the main concern, the image quality and realism is sometimes of higher importance, depending on the purpose of the application. Simple anonymization methods, like blur or pixelizing, conserve privacy and are enough when it comes to detection tasks \cite{hukkelaas2023does}. In contrast, getting the view direction or other keypoints of a pedestrian are lost using these approaches. In order to ensure no corruption of the data, we propose to anonymize the full body with a generated, highly detailed, one that is dissimilar to the original person whilst still retaining information, like skin color, anatomy, i.e. (see \cref{fig:title}). This is already done partially in DeepPrivacy2 \cite{hukkelaas2023deepprivacy2}, but with limited image resolution quality and without focus on retaining features.
Our main contributions are:
\begin{itemize}
    \item A novel pipeline that anonymizes people in arbitrary images for the use in neural network training, dataset creation and data storage (i.e. on a blackbox for vehicles)
    \item Evaluation that shows the impact of image anonymization on model training
    \item Use of replaceable pre-trained diffusion models for general anonymization purposes
\end{itemize}
This work is structured the following way. We first describe related work in the area of person anonymization and diffusion models (\cref{sec:related_work}). We then start to explain our method in detail in \cref{sec:main}, describing the different building blocks in detail. At the end in \cref{sec:results}, we show results of our method mainly compared to DeepPrivacy2 \cite{hukkelaas2023deepprivacy2}, a state of the art full body anonymization method. The main test cases are image quality, anonymization guarantee and YOLO \cite{redmon2016you,yolov8_ultralytics} training behaviour.

\section{Related Work}
\label{sec:related_work}
Image anonymization is widely used in practice, mainly preserving the privacy of people, but also for number plates and other sensible information. Some classic methods are the blur, pixelization or masking filter, which are easy to implement, but also degrade the image quality and corrupt important features \cite{hukkelaas2023does}. Recently, generative adverserial networks (GANs) were used to generate high quality realistic images with the capability to be used for image inpainting \cite{karras2020analyzing,ramesh2021zero}. Hukkelas et al. \cite{hukkelaas2023deepprivacy2} propose the use of multiple GANs for full body and face anonymization of people. They use pose estimation and continuous surface embeddings \cite{neverova2020continuous} to guide the generative network and get impressive results. However, their model outputs at $256 \times 256$ resolution which is enough for smaller images, but fails at the reconstruction of details, such as eyes and facial features, needed for tasks like intention recognition. It is shown that low resolution results impact the model performance when comparing against the original dataset \cite{hukkelaas2023does}.

\subsection{Diffusion Models}
Diffusion models (DMs) \cite{ho2020denoising,saharia2022palette} are an alternative to GANs for image generation. They operate on the basis of reversing a Markov chain.

\begin{figure}
    \centering
    \includegraphics[width=0.9\textwidth]{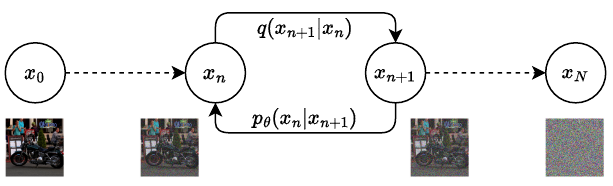}
    \caption{The Markov chain used for DMs. The initial state $x_0$ is corrupted by iteratively adding noise $q(x_{n+1} | x_n)$ until arriving at the fully noised image $x_N$.}
    \label{fig:diffusion_model_chain}
\end{figure}
The procedure starts with defining a scheduler which sets the noise added at each timestep $n \in [0, N]$ in the chain. The original image $x_0$ is then distorted by applying the distortion $q(x_{n+1} | x_n)$ defined by the scheduler iteratively. This is illustrated in \cref{fig:diffusion_model_chain}. During training, a single chain link is sampled and the model is trained to output the noise $p_\theta(x_n | x_{n+1})$. For image generation, $x_N$, a random normalized distributed image, is used as a starting point. To get a deterministic output, a seed can be used for the random number generator to generate consistent $x_N$. For inpainting, masking regions to not contain noise is possible, as well as starting at an arbitrary $x_n$ to corrupt the image less. Originally proposed as an unconditional generation model \cite{ho2020denoising}, conditional embeddings can also be added to the model, like text. This leads to the recently proposed Stable Diffusion \cite{Rombach_2022_CVPR,podell2023sdxl} for the possibility of context aware generation. The latest model is trained to output $1024 \times 1024$ images containing stunning detail and facial features. Additionally, using ControlNet \cite{zhang2023adding}, embeddings can be changed to match certain features, like preserving edges, image depth or poses for people.
Previously, face anonymization using DMs \cite{klemp2023ldfa} showed promising results. On the contrary, running diffusion models is much slower than GANs, because of the need of iteratively applying the model to solve the reverse markov chain. To speed up this process, Adversarial Diffusion Distillation \cite{sauer2023adversarial} is proposed, which reduces the execution time to a fraction of the original whilst still achieving high quality results.

\section{Anonymization Pipeline}
\label{sec:main}
In this section, we explain our anonymization pipeline in detail. We call our method FADM (\textbf{F}ull-Body \textbf{A}nonymization using  \textbf{D}iffusion  \textbf{M}odels) and mainly focus on full body people anonymization in this paper, but the pipeline proposed can also be adapted to different classes, which will be elaborated in the future. We start with a brief overview of the whole method and later discuss each building block in detail.

\begin{figure}
    \centering
    \includegraphics[width=0.9\textwidth]{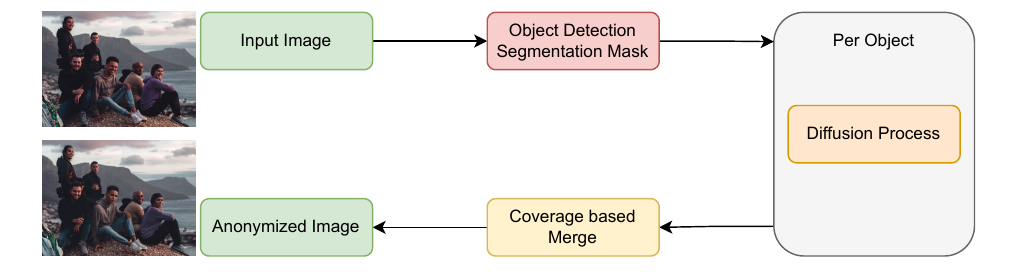}
    \caption{A high level overview of our anonymization pipeline.}
    \label{fig:anymous_pipeline}
\end{figure}

In \cref{fig:anymous_pipeline}, a high level overview of our pipeline is given. Object detection and instance segmentation is applied to the input image to get the bounding box and per-pixel segmentation mask of each object. For every instance, we dispatch a text-to-image diffusion model with a general prompt to inpaint the mask with plausible information. When using parallelization, the resulting list of cropped images need to be ordered back to front for compositing, which we do based on pixel coverage.

\subsection{Object Detection and Segmentation}
The first step in our pipeline is to detect the objects to anonymize. This is done by using a pre-trained YOLOv8 \cite{yolov8_ultralytics} detector by ultralytics (namely "yolov8\_m-seg" trained on COCO) paired with instance segmentation for masking objects. The result is a bounding box and instance mask for every object used for the per-object pipeline described in the next section.

\subsection{Diffusion Process}
\label{sec:diffusion_process}
After retrieving cropped images per object, we inpaint the segmentation mask using a text to image diffusion model. For the text prompt, we set \textit{"RAW photo, subject, 8k uhd, dslr, soft lighting, high quality, film grain, Fujifilm XT3"} as the positive prompt and \textit{"deformed iris, deformed pupils, semi-realistic, cgi, 3d, render, sketch, cartoon, drawing, anime), text, cropped, out of frame, worst quality, low quality, jpeg artifacts, ugly, duplicate, morbid, mutilated, extra fingers, mutated hands, poorly drawn hands, poorly drawn face, mutation, deformed, blurry, dehydrated, bad anatomy, bad proportions, extra limbs, cloned face, disfigured, gross proportions, malformed limbs, missing arms, missing legs, extra arms, extra legs, fused fingers, too many fingers, long neck"} as the negative one. A pre-trained Stable Diffusion (SD) model \cite{Rombach_2022_CVPR,huggingfacesd2_1,podell2023sdxl,huggingfacesdxl} is used as the generator, which leads to resolutions of $512 \times 512, 768 \times 768$ and $1024 \times 1024$ respectively, an increase of up to $16$ times the pixels compared to previous methods. SDXL is the most compute intensive workload. As a good fit for quality vs. performance, we recommend using Stable Diffusion 2.0 inpainting \cite{huggingfacesd2_0} at a resolution of $768$.

\begin{figure}
    \centering
    \includegraphics[width=0.95\textwidth]{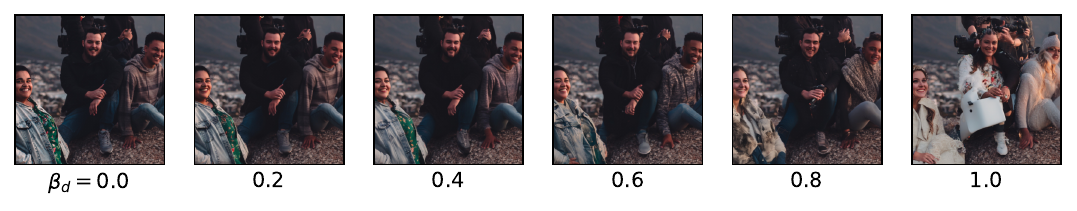}
    \caption{How the parameter $\beta_d$ influences the anonymization strength of a person instance.}
    \label{fig:dn_strength}
\end{figure}

In order to guide the diffusion process, we propose a noise value of maximum $\beta_d = 0.6$. This means, we do not take the full $N$ steps, but start at state $x_{\lfloor \beta_d N \rfloor}$ of the Markov chain. As a result, the original image is mixed into the noisy image by a factor of $1 - \beta_d$. Thus, we can say that $\beta_d$ modifies the amount of anonymization of the object. We visualized the influence of this parameter in figure \cref{fig:dn_strength}. A large value changes the image significantly, as most of the initial image is noise and  has no information about the original person. We also see clearly how lower values of $\beta_d$ lead to less anonymization.

\subsection{Coverage based Merge}
Depending on the implementation of the per-object process, the resulting images are unordered. This means that we do not know the back to front order of all objects, but this is of essential importance, as objects in the foreground can still contain some parts of background objects and vice versa. When doing everything iteratively, this does not pose a problem, as for every anonynmization, the previous one is finished and already merged into the final image.

In order to improve performance, we use parallel processing for precomputing the cropped images before executing the diffusion process. This has the benefit of being able to batch the per object images together in a single call to the diffusion model if enough memory is availbale on the gpu. Images generated that way always contain the original background pixel data for every object. To reduce overlapping artifacts and get a back to front ordering, we sum up the instance segmentation mask values of each object, which effectively gives us the coverage in pixels. We assume that objects in the foreground are larger and occupy more space than objects in the background. Then, we sort the results upwards based on the coverage values and merge the images together to get the final anonymized result. This method is proposed by Hukkelas et al. \cite{hukkelaas2023deepprivacy2} as \textit{Recursive Stitching}.

An alternative to coverage based sorting is using depth estimation \cite{ranftl2022midas,bhat2022zoedepth,depthanything} and ordering the image crops from back to front. Although we have not tested this approach and do not explain more details about it in this paper, we think it is interesting to look at in the future. The reason is that coverage based merging assumes objects in the foreground to be larger than ones in the background. In general, this is not the case, especially when children are in the image that are smaller than adults. Correct per-pixel depth estimation has no such assumptions. In contrast to depth estimation, expanding our method to different object classes poses a problem for coverage based merging, as object scales can vary drastically then.

\section{Results}
\label{sec:results}
In this section, we show results of the proposed pipeline. We start with assessing the general image quality, then show if anonynmization has an impact when training a YOLOv8 \cite{yolov8_ultralytics} object detector, as well as a Mask2Former \cite{cheng2021mask2former} segmentation model. Last but not least, we explain that our method ensures anonymization by testing against re-identification algorithms on the market1501 and LAST datasets. Additionally, we also perform a face only re-identification evaluation.

\subsection{Image Quality}
A very important part of anonymization is preserving image quality. We already see some examples in \cref{fig:title}, where our method is used to anonymize high resolution images from different scenarios. Previous methods, like DeepPrivacy2 \cite{hukkelaas2023deepprivacy2}, are limited to $256 \times 256$ output resolution, which is way too low for high resolution images everywhere today. Our method can generate high quality images up to $1024 \times 1024$ resolution, greatly improving sharpness of the anonymized image part. 

\begin{figure}
    \centering
    \includegraphics[width=0.7\textwidth]{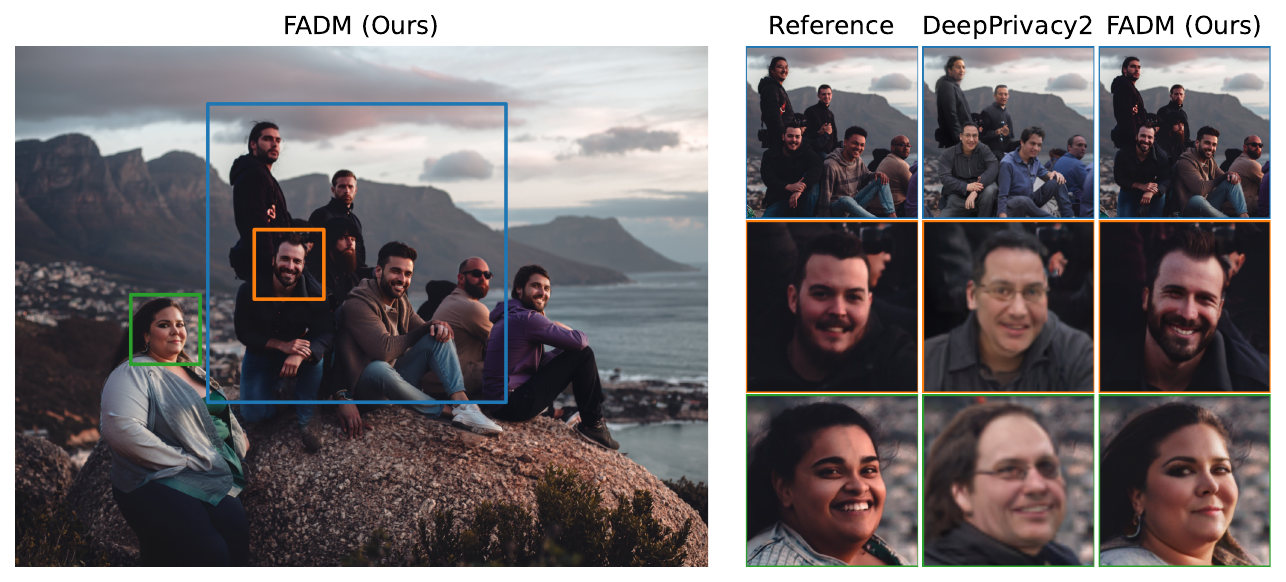}
    \caption{Comparison of DeepPrivacy2 and our method on a high resolution stock photo \cite{witbooi2023}.}
    \label{fig:plot_img_chad}
\end{figure}

\begin{figure}
    \centering
    \includegraphics[width=0.3\textwidth]{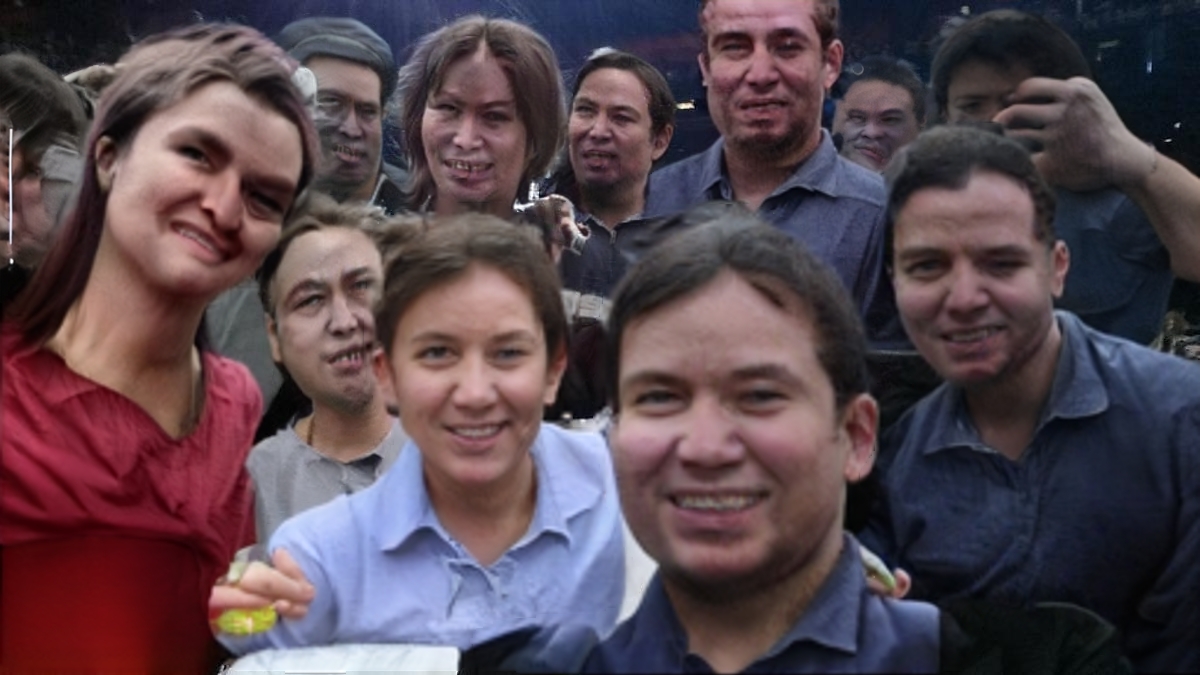}
    \includegraphics[width=0.3\textwidth]{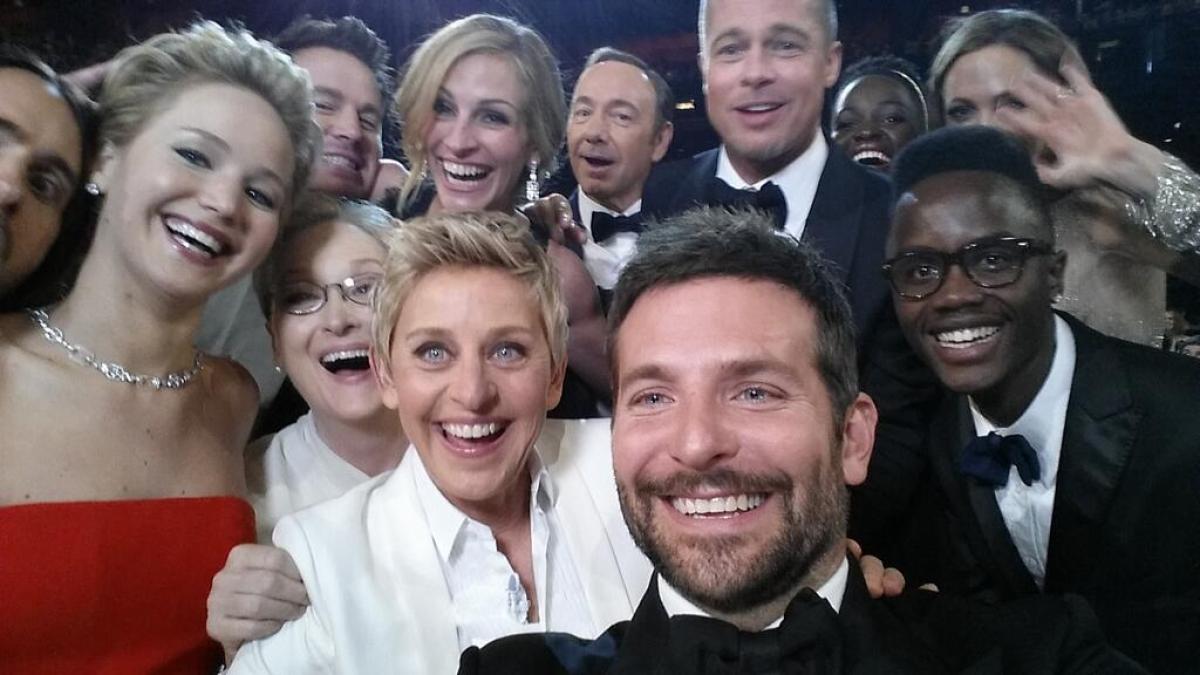}
    \includegraphics[width=0.3\textwidth]{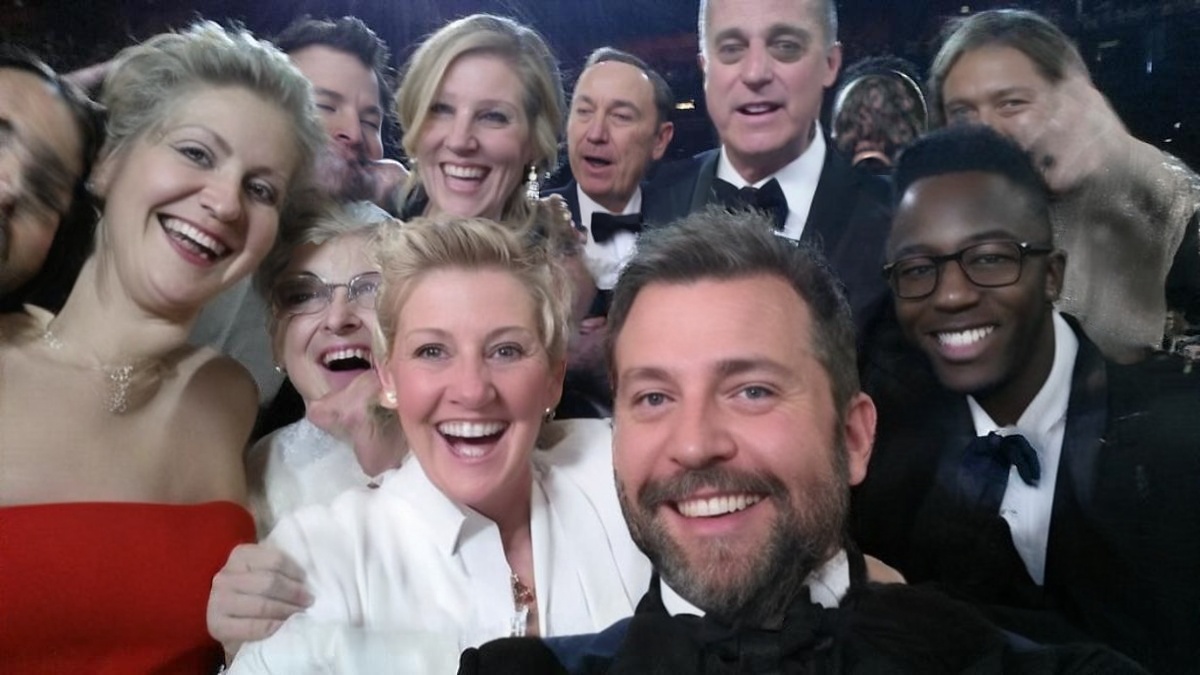}
    \caption{Comparison of anonymizing Ellen's Oscar selfie (middle) with DP2 (left) and our method (right).}
    \label{fig:ellen_selfie}
\end{figure}

However, resolution is not everything, the context and overall look of an image, like lighting, shadows and features are also very important. \cref{fig:plot_img_chad} compares DeepPrivacy2 (DP2) with our method on an image of a group of people. Looking at the orange and green part, we clearly see the improvement in sharpness. Additionally, our method preserves the general image feel, in this case a warmer look, correct skin tone and shadows. The image from DP2 looks more out of context and not very well integrated into the original image with respect to lighting and compositing. This behaviour is seen in many images, like \cref{fig:ellen_selfie}, where we anonymize a group of people. Our method creates a much more realistic look than DP2. Especially when looking at faces in that image, as well as the clothes, which look more natural for the context of that scene (shot at the Oscars).

\begin{figure}
    \centering
    \includegraphics[width=0.7\textwidth]{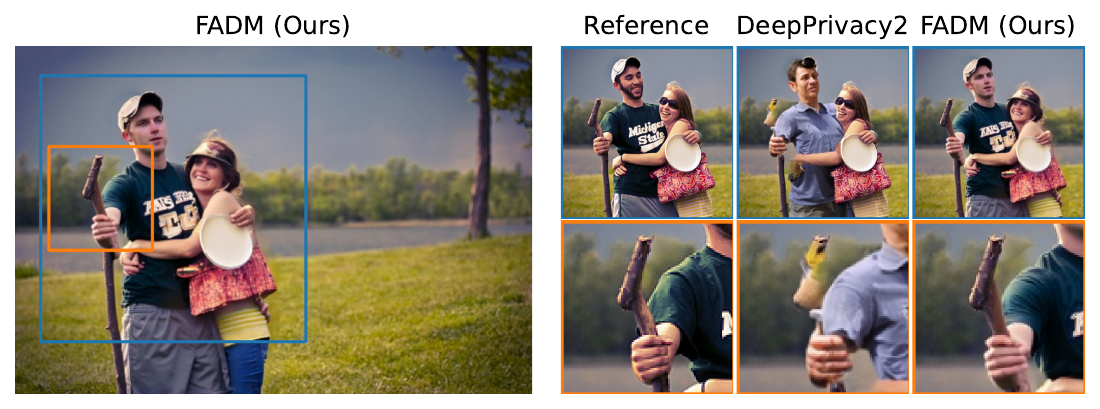}
    \caption{Comparison of DeepPrivacy2 and our method on a low resolution image from the COCO Dataset \cite{lin2014microsoft}.}
    \label{fig:plot_img_1569}
\end{figure}

When looking at lower resolution images, we still see the same behaviour, like in \cref{fig:plot_img_1569}. The composition of DP2 is better than in \cref{fig:plot_img_chad}, but still lacks detail. For example, the wooden stick is corrupted in the DP2 image, whilst realistic looking when using our method. Also, the face of the woman is ignored in DP2, but that may be a problem with the detector the algorithm uses. Our method not only anonymizes the person, but also clothes in context. As text is problematic for diffusion models (see also \cref{fig:title}), the logo on the shirt is anonymized, but mostly contains gibberish text. However, this is only a minor problem as it retains the context of \textit{a shirt containing a logo}.

\begin{table}[t]
    \centering
    \small
    \begin{tabular}{l C{2cm} C{2cm}}
        \toprule
         Method & IS $\uparrow$ & FID $\downarrow$ \\
         \midrule
         Reference & \textbf{27.39} & - \\
         Blurring & 20.83 & 31.87 \\
         Masking & 23.15 & 22.99 \\
         Pixelization & 20.39 & 36.39\\
         DeepPrivacy2 & 25.01 & 6.03 \\
         \textbf{FADM (Ours)} & 27.23 & \textbf{3.02} \\
         \bottomrule
    \end{tabular}
    \caption{Inception Score (IS) and Frechet Inception Distance (FID) of different anonymization methods on a subset of the COCO Dataset \cite{lin2014microsoft}.}
    \label{tab:scores}
\end{table}

To verify our image quality findings, we anonymized a subset of the COCO dataset containing 10k images from different situations. The resulting images are then used to calculate the Inception Score (IS) \cite{salimans2016improved} and Frechet Inception Distance (FID) \cite{heusel2017gans,chong2020effectively}, shown in \cref{tab:scores}. These metrics are commonly used for generative AI models to calculate the quality of the output. A higher IS corresponds to better image quality where as a lower FID is better in general. FID is measured relative to the original dataset, thus there is no score for the reference. We see that classical methods, like blurring, masking and pixelation, achieve the lowest quality possible. DP2 improves the IS and FID drastically compared to the previously mentioned algorithms. Our method outperforms previous algorithms by nearly being tied with the reference dataset for the IS and halves the FID compared to DP2.

\subsection{Object Detection}
We already showed that our method significantly improves the image quality compared to previous methods. The next question is: "Does it even matter for training AI models?" There are many different problems in the computer vision space. We cannot test all of them in this paper. We focus on object detection, as it is an important task and a good indicator if anonymization impacts training performance \cite{hukkelaas2023does}.

\begin{figure}
    \centering
    \includegraphics[width=0.66\textwidth]{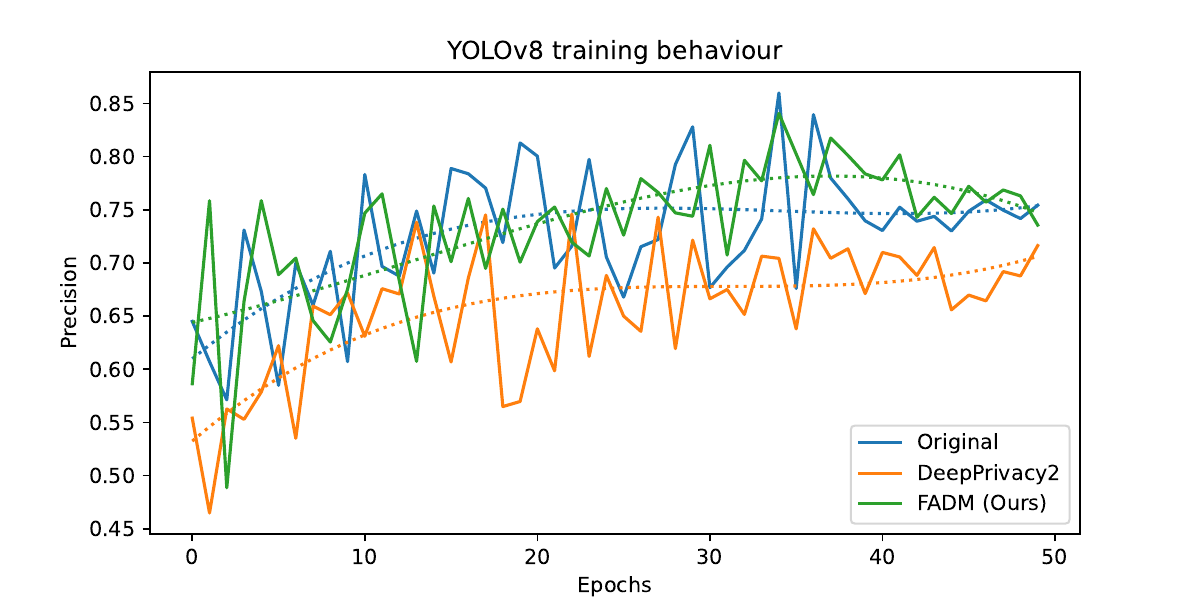}
    \includegraphics[width=0.33\textwidth]{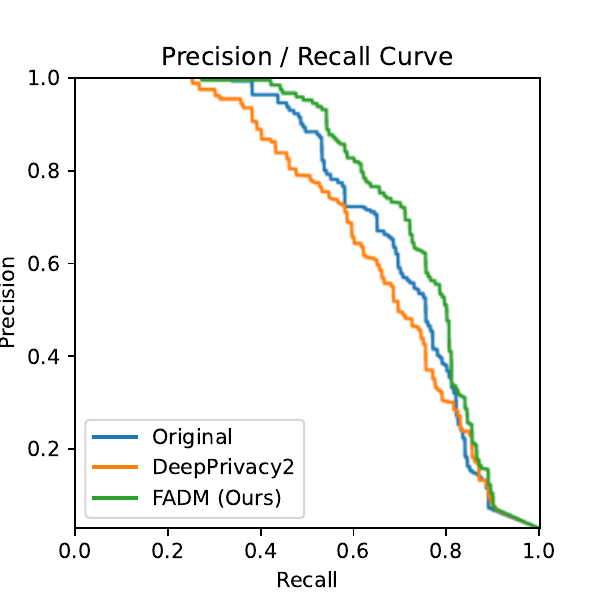}
    \caption{Training behaviour of a YOLOv8 models on the left and their corresponding Precision Recall curve on the right. Training uses the original dataset (blue), anonymized one with DP2 (orange) and with our method (green).}
    \label{fig:plot_training_behaviour}
\end{figure}

As a first experiment we trained a YOLOv8 model multiple times on a subset of the COCO dataset \cite{lin2014microsoft} to detect people and tested the result on a static validation set of real images. The first training was done with the reference dataset, the second one used anonymization by DP2 and the third one used our method. In \cref{fig:plot_training_behaviour}, we plot the precision of the model on the validation set for the first 50 epochs. We clearly see that training with DP2 results in a worse precision than using the original dataset. In contrast, using our method, the model performance is not degraded compared to the original one, showing that anonymization has no impact on people detection in this task.


This is supported by the precision-recall curve in \cref{fig:plot_training_behaviour}, which shows the performance of the best model trained on the three datasets mentioned earlier. A better result is indicated by a curve that fits closer to the top right corner. When using DP2, the model underperforms compared to the original dataset, while our method shows a slight improvement over the original. There is no clear explanation for this behaviour. However, it is possible that our algorithm produces sharper edges and generally clearer images than the original dataset. For instance, slight blur, i.e. depth of field, is mostly eliminated, resulting in a sharper and clearer image of the generated person. Although this would result in the model training only on sharp images, more blurry ones are still anonymized to be blurred.

\begin{table}[]
    \centering
    \begin{small}
    \begin{tabular}{lccccc}
        \toprule
        Anon. method & $\text{IoU}_{\text{Person}} \uparrow$ & $\Delta\text{IoU}_{\text{Person}}^{\text{rel}}$ & $\text{IoU}_{\text{Rider}} \uparrow$ & $\Delta\text{IoU}_{\text{Rider}}^{\text{rel}}$ &    $\text{IoU}_{\text{Human}} \uparrow$  \\
        \midrule
        Baseline & 0.836 & 0.00\% & 0.443 & 0.00\% & 0.894 \\
        \midrule
        Naive & & & & &\\
        \midrule
        \textsc{Blurring} & \textbf{0.898} & \textbf{+7.4\%} & \textbf{0.69} & \textbf{+55.75\%} & \textbf{0.94} \\
        \textsc{Masking} & 0.006 & -99.2\%	& 0.070 & -84.2\%	&0.018	\\
        \textsc{Pixelization} & 0.247 & -70.45\%			& 0.142	& -67.94\%		& 0.263 \\
        \midrule
        Deep learning-based &  &  &  &  &  \\
        \midrule
        \textsc{DeepPrivacy2} & 0.624  	& -25.35\%		& 0.587	& +32.5\%	& 0.663\\
        \textbf{\textsc{FADM} (ours)} & 0.770 	& -7.89\%	& 0.562	& +26.86\%	& 0.776\\
        \bottomrule
    \end{tabular}
    \caption{Impacts of anonymization methods on Mask2Former semantic segmentation}
    \end{small}
    \label{tab:mask2former}
\end{table}

Secondly, we show the impact of all anonymization methods already mentioned in \cref{tab:scores}. For this, we trained a Mask2Former \cite{cheng2021mask2former} on the cityscapes dataset \cite{cordts2016cityscapes} as a baseline. To evaluate the impact of the anonymization methods on the model training, we anonymized the training split and evaluated the model on the raw images of the validation split. For the training setup and evaluation metrics, we follow the setup used in \cite{klemp2023ldfa}. The results are shown in \cref{tab:mask2former}. We see that our method outperforms DP2, masking and pixelization regarding the IoU of people whilst falling slightly behind in the "rider" class. Interestingly, blurring the mask outperforms all other methods and improves the model performance compared to the baseline.

\subsection{Re-Identification}
So far, we have only shown that our method produces realistic images and does not sacrifice model performance when it comes to object detection. However, this paper is designated to anonynmization and we have to validate that the proposed method actually anonymizes people. To measure anonymization, we opted to use a person re-identification model as a metric. Keep in mind, this is in no means a proof that there exists no algorithm that can defeat our anonymization method.

\begin{table}[]
    \centering
    \small
    \begin{tabular}{lC{2cm}C{2cm}}
        \toprule
         Method & Rank-1 $\downarrow$ & mAP $\downarrow$ \\
         \midrule
         Original & 94.4\% & 82.6\% \\
         Blurring & 13.1\% & 11.0\% \\
         Masking & 12.7\% & 10.4\% \\
         Pixelization & \textbf{12.3}\% & \textbf{10.3}\% \\
         DeepPrivacy2 & 40.8\% & 34.9\% \\
         \textbf{FADM (Ours)} & 67.8\% & 53.7\% \\
         \bottomrule
    \end{tabular}
    \caption{The rank-1 accuracy and mean average precision (mAP) of OSNet on the Market1501 dataset when using different anonymization methods.}
    \label{tab:reid_m1501}
\end{table}

As a re-identification method, we use OSNet \cite{zhou2019osnet,zhou2021osnet} via the torchreid \cite{torchreid} library and anonymize the Market1501 \cite{zheng2015scalable} dataset. OSNet is trained on the dataset, which contains 1501 identities in 32k images. The task of OSNet is then to match a given image with a target identity from the dataset. Our test used the anonymization method to anonymize the person before OSNet is applied and a method succeeds when the model outputs the wrong identity.

The result is described in \cref{tab:reid_m1501}, where the Rank-1 accuracy and mean average precision (mAP) is shown. Rank-1 accuracy describes the number of identities correctly matched by the algorithm. Mean average precision is based on multiple metrics and mainly calculated over recall values. We see that OSNet does a great job at identifying people in the original dataset. DP2 anonymizes the images and yields a much lower accuracy for the algorithm, showing its anonymization capabilities. In contrast to the original paper \cite{hukkelaas2023deepprivacy2}, we get worse results in this test with the code provided by the authors. They original reported a Rank-1 accuracy of 44.7\% and mAP of 8.5\%. Our method is not as good as DP2 when it comes to anonymization in this test, but still reduces the overall score of the OSNet re-identification method significantly. In general, we can say that our method trades anonymization for a large increase in image quality. Of course, destroying an image using masking, bluring or pixelization would yield a high anonymization at the cost of image quality.

Please note that our method is designed for high resolution image inpainting. The dataset used for testing, Market1501, has a resolution of $128 \times 256$, which is very low to start with.

\begin{figure}[]
    \centering
    \scriptsize
    \includegraphics[width=0.6\textwidth]{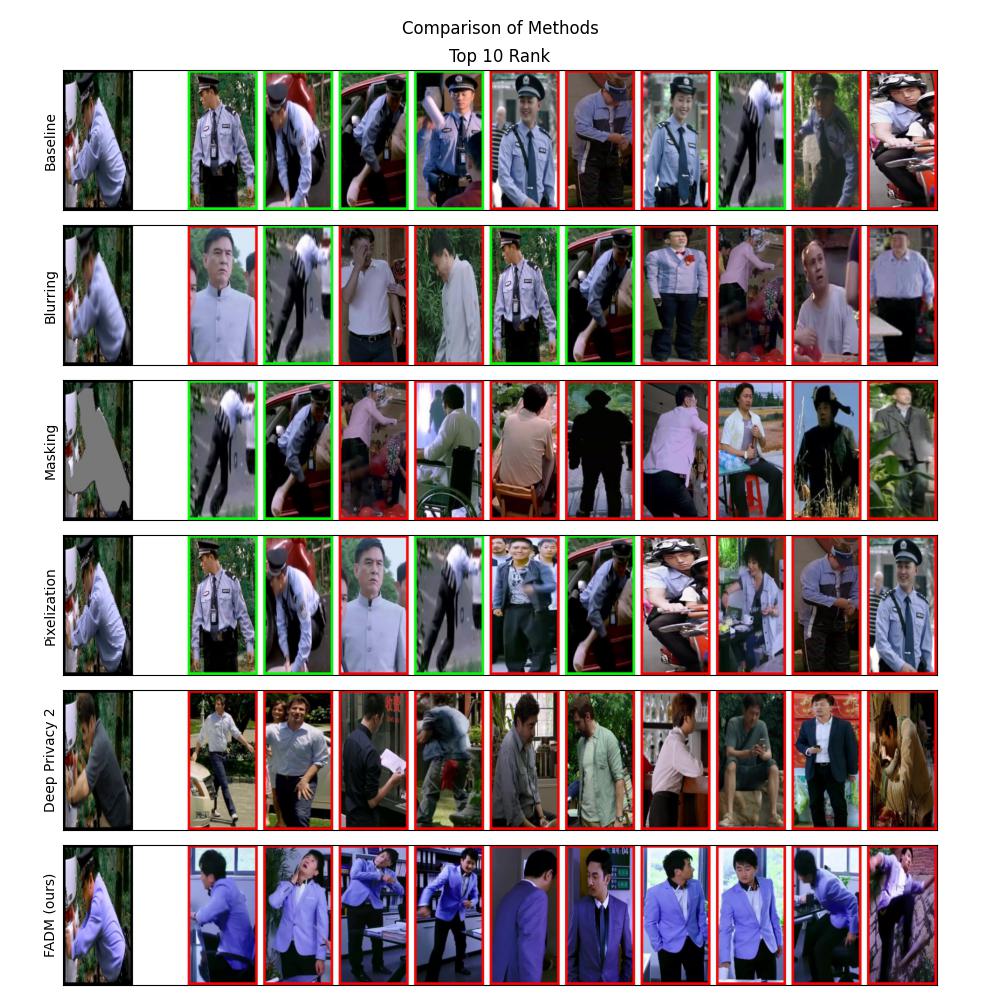}
    \caption{An example from the LaST dataset \cite{shu2021large} anonymized using our method and tested by the OSNet re-identification algorithm. It visualizes the anonymized query image on the left and the positive (green) and negative (red) matches from the gallery on the right.}
    \label{fig:plot_rank10}
\end{figure}

Additionally, OSNet is trained with only 1501 identities, so there is the possibility that many generated identities map to a close latent space representation.
That's why we also tested our method on the LaST dataset \cite{shu2021large}, which contains over 10000 identities in around 228000 images.
The results are shown in \cref{tab:last_reid}.

\begin{table}[]
    \centering
    \small
    \begin{tabular}{lC{2cm}C{2cm}C{2cm}}
        \toprule
        Method & Rank-1 $\downarrow$ & Rank-10 $\downarrow$ & mAP $\downarrow$ \\
        \midrule
        Original & 69.6\% & 86.0\% & 25.6\% \\
        Blurring & 21.6\% & 41.8\% & 5.9\% \\
        Masking & 4.5\% & 13.2\% & 1.4\% \\
        Pixelation & 44.7\% & 67.2\% & 13.3\% \\
        DeepPrivacy2 & \textbf{3.4\%} & \textbf{8.8\%} & \textbf{1.1\%} \\
        \textbf{FADM (Ours)} & 41.6\% & 65.4\% & 12.7\% \\
        \bottomrule
    \end{tabular}
    \caption{The rank-1 accuracy and mean average precision (mAP) of OSNet on the LaST dataset when using different anonymization methods.}
    \label{tab:last_reid}
\end{table}

\cref{fig:plot_rank10} shows an example of the rank 10 retrieval. The query image is anonymized using our method and the positive and negative matches are shown on the right. 
This sample shows that re-identification is very dependent on the color distribution on the image. 
As soon as the color of the clothes changes, as in the image anonymized by DeepPrivacy2 and FADM, the main clothing in the images retrieved from the gallery match the query images.
Even the slighter color change in the FADM image is enough to fool the re-identification algorithm.

\begin{table}[]
    \centering
    \begin{tabular}{lc}
        \toprule
        Method & Rank-1 accuracy \\
        \midrule
        Reference &  89.41\% \\
        Blur &  12.26\% \\
        Masking &  0.00\% \\
        Pixel &  4.18\% \\
        Ours &  \textbf{2.78\%} \\
        \bottomrule
    \end{tabular}
    \caption{The rank-1 accuracy of the face re-identification algorithm on the faces in the wild dataset anonymized with different algorithms.}
    \label{tab:faceid}
\end{table}

\begin{figure}[]
    \centering
    \includegraphics[width=0.7\textwidth]{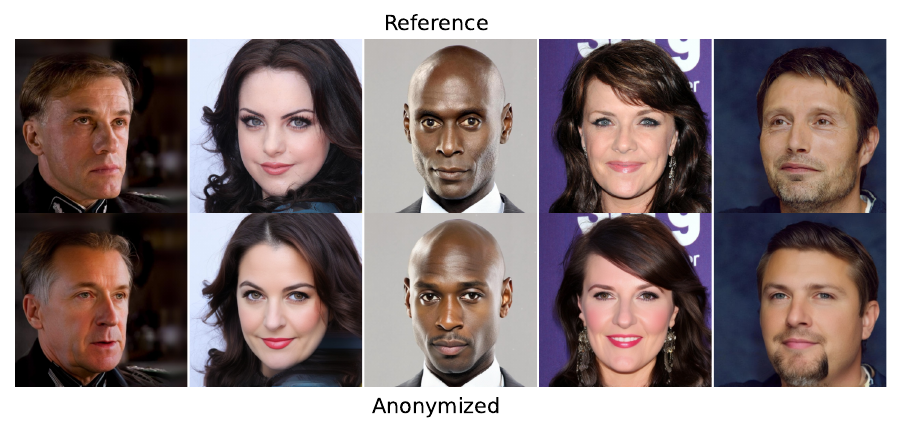}
    \caption{Some example images from the CelebAMask-HQ dataset \cite{CelebAMask-HQ} anonymized using our method and tested by the face re-identification algorithm.}
    \label{fig:plot_faceid}
\end{figure}

We also tested face re-identification on higher resolution images using the \textit{face\_recognition} library \cite{ageitgey2016facerecognition}, which uses the dlib face recognition pipeline \cite{king2009dlib} achieving 99.38\% accuracy on the labeled faces in the wild dataset (LFW) \cite{LFWTech,LFWTechUpdate}. \cref{fig:plot_faceid} shows a few images used for testing while Tab. \ref{tab:faceid} shows the performance of the face recognition model when used on both datasets. 

It is clear that our method sucessfully anoynmizes identities in the dataset and deceives the re-identification algorithm with a much lower accuracy than with the original dataset.

\subsection{Limitations}
\label{sec:problems}
In this section, we want to show some limitations of our approach. Our method highly depends on the generation quality of the diffusion model. Although current models produce photorealistic, high quality images, some cases exist where the output is corrupted.

\begin{figure}
    \centering
    \includegraphics[width=0.5\textwidth]{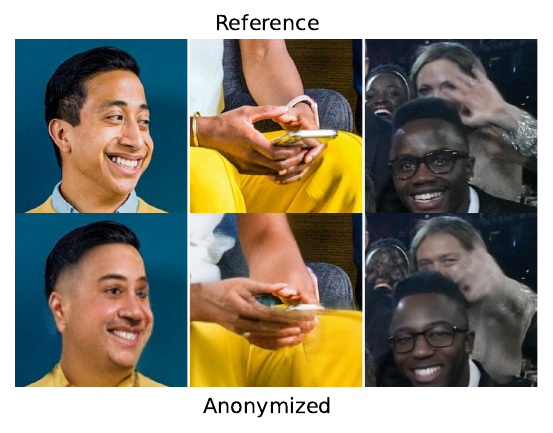}
    \caption{Failure cases where the diffusion model generated deformed hands and faces when applied to images from the CelebAMask-HQ \cite{CelebAMask-HQ}.}
    \label{fig:obama_problems}
\end{figure}

We show some problems in \cref{fig:obama_problems}. The face looks slightly deformed and the right eye is not very well reconstructed. We think this happens due to the model being trained on less data for certain angles of human faces as well as the general problem of generating faces at lower resolutions. In this example, the model had to generate a complex pose of a whole person. The second example shows a problem reconstructing hands, which is very common when using the latest diffusion models and may be improved in future versions. The third one shows that faces are sometimes completely removed, as the one of the woman in the background. Additionally, the hand gets blurred and does not look like a hand anymore.

These scenarios do not occur as often as when generating an image from scratch, i.e. from full noise, but can occasionally occur. However, our method is still valid as the generator can be replaced by any other generative model. We think with research advances in the next years, the hand problem of generative models can be reduced or eliminated and then directly used with our approach.

There is also no support for video streams yet. Though the pipeline is capable of working on image sequences, Stable Diffusion is not capable of generating temporally consistent image sequences out of the box. At the end of doing the research for this paper, Stable Video Diffusion \cite{blattmann2023stable} was proposed, showing significant improvements in temporal stability that should be easily integratable into our pipeline.

To reproduce our results, we refer to the publicly available source code of our project \cite{sourceGit}.

\section{Conclusion}

In this work, we proposed a novel pipeline for full-body people anonymization. Our method achieves higher image quality than previous methods regarding resolution, realism and image composition. We provided justification that the proposed method anonymizes people by using re-identification algorithms and deceiving them. Using an object detector trained on our anonymized dataset, we showed that the resulting model is on par and sometimes outperforms a model trained on the original dataset. This is in contrast to the results of previous method \cite{hukkelaas2023does}, where image quality and resolution are worse. The proposed pipeline can be used with any text-to-image model and thus can be updated with the latest models in the future. This is important as new methods, like Adverserial Diffusion Destillation \cite{sauer2023adversarial}, improve the speed and quality of diffusion models and thus our pipeline. For future research, combining a prompt sampler with the diffusion model seems interesting to improve variety and improve anonymization strength. Furthermore, the proposed method is only executed on single images, making it unsuitable for videos as the diffusion model relies heavily on the initial noise. Integrating a temporal step combined with Stable Video Diffusion \cite{blattmann2023stable} can lead to usable video anonymization.

\subsection*{Honorary Mentions}
This paper emerged during the research projects \textit{ANYMOS - Competence Cluster Anonymization for networked mobility systems} and just better DATA (jbDATA) supported by the German Federal Ministry for Economic Affairs and Climate Action of Germany (BMWK) and was founded by the German Federal Ministry of Education and Research (BMBF) as part of \textit{NextGenerationEU} of the European Union.

\centering
\raisebox{0.6\height}{\includegraphics[width=0.4\textwidth]{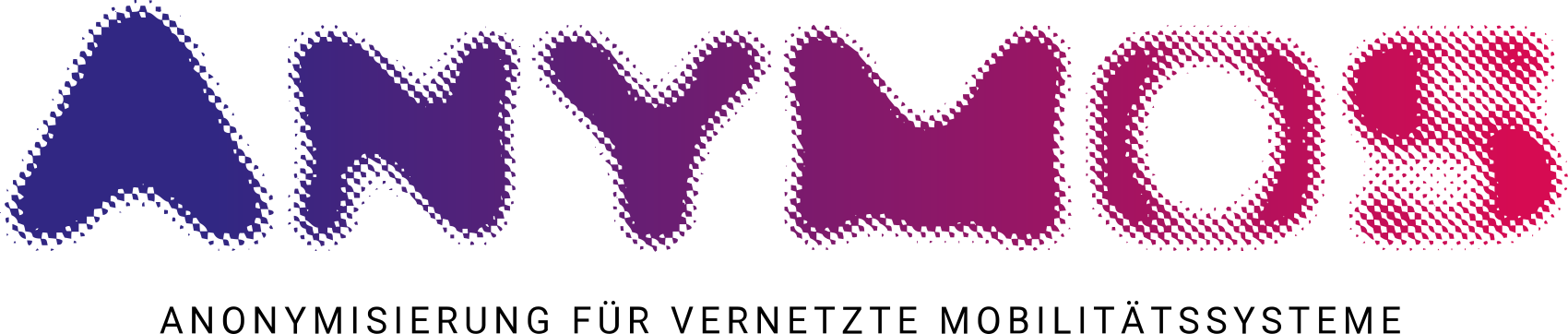}}
\includegraphics[width=0.4\textwidth]{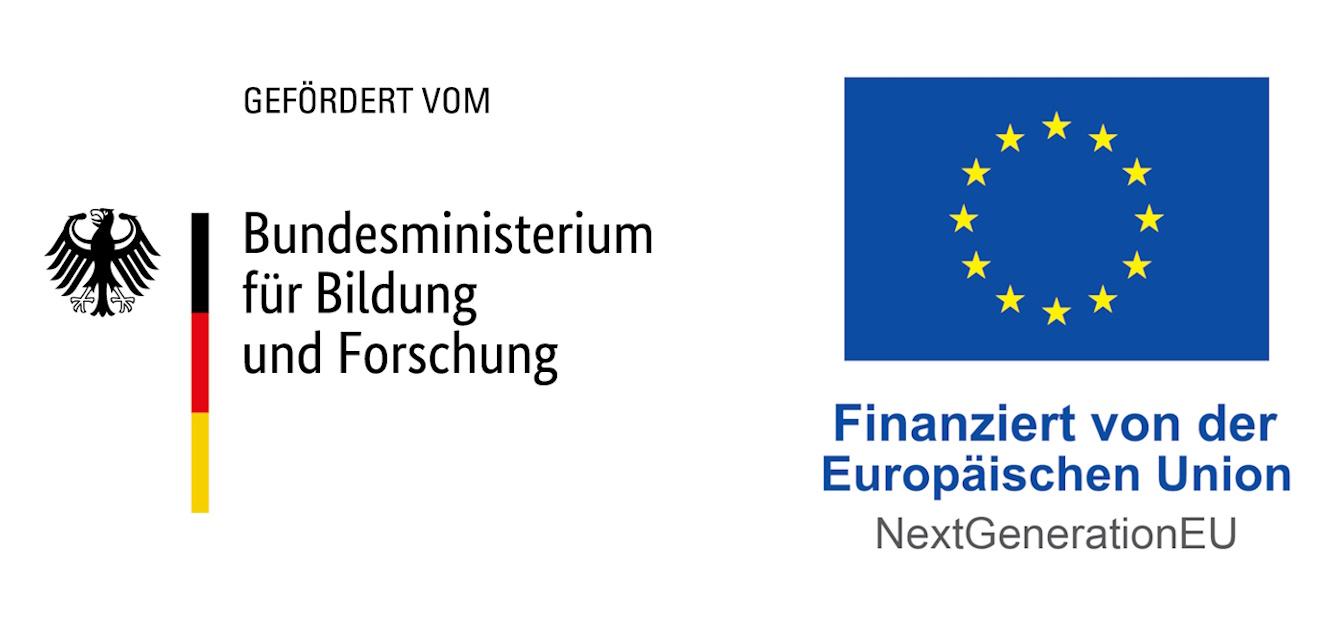}

%
%
\bibliographystyle{plain}
\bibliography{main}
\end{document}